\documentclass{article}
\pdfoutput=1

\PassOptionsToPackage{numbers, compress}{natbib}
\usepackage{natbib}
\usepackage[final]{neurips_2022}

\usepackage[utf8]{inputenc} 
\usepackage[T1]{fontenc}    
\usepackage{url}            
\usepackage{booktabs}       
\usepackage{amsfonts}       
\usepackage{nicefrac}       
\usepackage{microtype}      
\usepackage{xcolor}         
\usepackage{algorithm}
\usepackage{algpseudocode}
\usepackage{bxcjkjatype}
\usepackage[linguistics]{forest}
\usepackage{tabularx,ragged2e}
\usepackage{graphicx}
\usepackage{amsmath,amssymb,amsfonts}
\usepackage{caption}
\usepackage{multirow}
\usepackage{subcaption}
\title{EuclidNet: Deep Visual Reasoning for Constructible Problems in Geometry}

\author{%
  Man Fai Wong \\
  City University of Hong Kong\\
  \texttt{mfwong29-c@my.cityu.edu.hk} \\
   \And
  Xintong Qi\\
  Columbia University\\
  \texttt{xq2224@columbia.edu} \\
  \And
  Chee Wei Tan\\
  	Nanyang Technological University\\
  \texttt{cheewei.tan@ntu.edu.sg} \\
}

\begin{document}

\maketitle

\begin{abstract}
  
In this paper, we present a deep learning-based framework for solving geometric construction problems through visual reasoning, which is useful for automated geometry theorem proving. Constructible problems in geometry often ask for the sequence of straightedge-and-compass constructions to construct a given goal given some initial setup. Our EuclidNet framework leverages the neural network architecture Mask R-CNN to extract the visual features from the initial setup and goal configuration with extra points of intersection, and then generate possible construction steps as intermediary data models that are used as feedback in the training process for further refinement of the construction step sequence. This process is repeated recursively until either a solution is found, in which case we backtrack the path for a step-by-step construction guide, or the problem is identified as unsolvable. Our EuclidNet framework is validated on complex Japanese Sangaku geometry problems, demonstrating its capacity to leverage backtracking for deep visual reasoning of challenging problems.
\end{abstract}

\section{Introduction}
\label{introduction}
%
Henri Poincaré once remarked ``{\it Geometry is the art of correct reasoning from incorrectly drawn figures}". Thus, reasoning in geometry often means drawing upon visual figures in a creative manner with {\it abundant trial and error}. This is true especially for geometric constructible problems that analyze the drawing of Euclidean shapes with straightedges and compasses \cite{hilbert1902foundations,pedoe}. Every constructible problem requires first a feasibility answer and then the exact sequence of straightedge-and-compass construction steps if feasible. For example, though Gauss proved in 1796 the constructibility of the regular seventeen-sided polygon, its explicit construction remains elusive until decades later as shown by J. Erchinger and H. W. Richmond \cite{pedoe}. Gelernter's geometry machine in 1959 would need a diagram computer to generate figures to validate its automated geometry theorem proving \cite{gelernter1958intelligent}. 

Visualization of figures is undeniably helpful to humans in developing intuition and grasping mathematical concepts in geometry. This process is not only perceptual but includes a visual validation step by step using trial and error to sieve through the visual information \cite{dreyfus1991status}. Kim in \cite{kim1989} proposed visual reasoning in geometry by drawing an analogy to human reasoning in identifying visual patterns and analyzing geometric feature information repeatedly. Other related works include the diagram reasoning in \cite{koedinger1990abstract}, program synthesis technique in \cite{gulwani2011synthesizing} to synthesize geometric constructions, the alignment of visual and textual cues in geometrical diagrams \cite{seo2015solving} and geometric deep learning \cite{geometricdeeplearning,naturemathai}. It has become increasingly important to understand how to leverage the availability of large data sets of visual figures to refine and accelerate the trial and error process in {\it automated visual reasoning}.

In this paper, we propose EuclidNet, a deep learning-driven framework that utilizes neural network-empowered visual reasoning techniques for constructible problems in geometry. In particular, the EuclidNet framework leverages the neural network Mask R-CNN to extract visual features from the images and categorize these visual features into points, lines, and circles. Since intersections may also possess significantly important geometric features,  we also implement another Mask R-CNN to extract intersections from the image and count them as points. Together, these features form a space where feasible construction steps are derived and form a rooted tree to traverse all possible features. From a variety of features in the tree, EuclidNet selects one geometrical feature to construct and then adds the construction step to the existing construction. Repeating the process, EuclidNet proposes sequences of constructions where backtracking is conducted once an existing construction is identical to the original input image, and a step-by-step reconstruction geometrical solution can thus be derived from the process. This procedure of divide-and-conquer and backtracking is reminiscent of Wang's algorithm for automated theorem proving of propositional logic \cite{wang1990computer,wangibm,wangacm}, where a statement is decomposed into multiple premises that are considered true, and the automated program exhausts the solution space to determine if the conclusion is valid. 

In summary, the contributions of our work are as follows.
\begin{itemize}
    \item We offer a new perspective on geometric construction using visual reasoning as a comprehensive procedure to discover solutions. Visual reasoning leverages the feature representation of geometric patterns extracted by Mask R-CNN to propose new construction possibilities.
    \item We introduce backtracking as a means of step-by-step construction solution finding. Once the existing construction is identical to the original input image, a solution is found, and EuclidNet backtracks the construction sequence to reproduce a {\it proof} of construction.
    \item We illustrate EuclidNet's relative performance against other approaches to solving geometric construction problems. We demonstrate the effectiveness of our proposed framework for complex geometric construction problems such as Japanese Sangaku geometry \cite{pedoesangaku}.
\end{itemize}

\section{Methodologies}

\subsection{Elementary Euclidean Constructions Procedure}
In a Euclidean plane, points, lines and circles are considered the fundamental constructing elements, which are also known as primitives. When we solve the Euclidean geometric construction problems, it is assumed that specific points are known as priori in the infinite Euclidean plane \cite{geretschlager1995euclidean}. With line and circle tools, only limited moves can be made with the existing points in the Euclidean plane:
\begin{enumerate}
    \item Given two non-identical points $A(x_1,y_1)$ and $B(x_2,y_2)$, we can draw the unique straight line $L=AA$ with the straightedge containing both points. The ray-line will be projected to the canvas boundary by the slope $m = \frac{y_2-y_1}{x_2-x_1}$ and y-intercept $b=y_1-mx_1$ or $b=y_2-mx_2$.
    \item Given two points $C(h,k)$, $M$ and $|CM| > 0$, we can use the compass to draw a unique circle $c = \{C;|CM|\}$ with $C$ being the center, $|CM|$ being the radii, and $M$ on the circle. 
\end{enumerate}
For each constructible problem, when new lines or circles are created on a diagram, EuclidNet aims to localize the intersections in each diagram from previous moves to create more points so as to explore other possible new constructions. Once we have exhaustively examined all possible actions in the given construction using the rules above, we are able to build the solution by considering all possible moves with a backtracking algorithm. This procedure is elaborated in Appendix \ref{appendixa2}. 

\subsection{Deep Visual Reasoning}
Although visual reasoning is second nature to humans given the visual inputs \cite{kim1989}, it is not trivial for computers to emulate this. Computers need to first sieve through geometric patterns that are visually depicted, and then comprehend logical relationships between patterns (e.g., isometries) in order to reason. Our approach is to employ deep learning in \cite{lecun2015deep} for geometric primitive extraction as well as image segmentation for pattern-seeking so as to enable data-driven pattern recognition for geometrical diagrams. In some sense, the interpretability of the trained machine learning model for automated visual reasoning is to turn geometrical diagrams into {\it proof without words} for human understanding. In EuclidNet, we implement our deep visual reasoning functionality with Mask R-CNN.
\begin{figure}[!h]
    \centering
    \includegraphics[width=\linewidth]{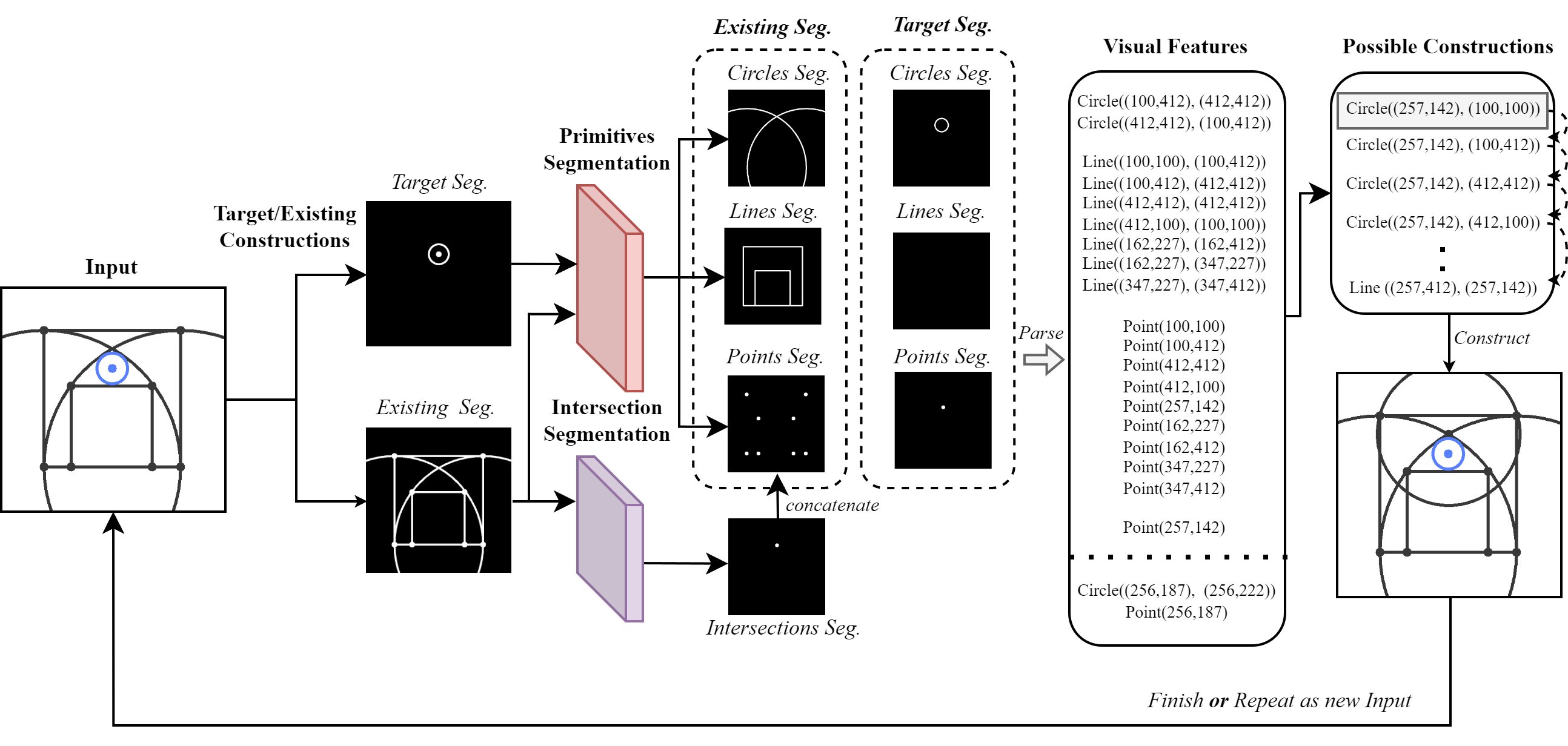}
    \caption{The detailed architecture of the proposed framework (EuclidNet) for solving geometric construction problems leverages visual reasoning and the backtracking algorithm. EuclidNet separates the image as target (in blue) and existing (in black) construction for an input image. EuclidNet consists of two pre-trained segmentation and localization models for geometric primitives and intersections respectively. The segmentation model is based on Mask R-CNN, which processes the input image and performs instance segmentation on the target and existing constructions segments. These pre-trained models receive the images of both the segments and the outputs. The outputs will then be parsed as visual features with their primitive type and location plotted on canvas. All permutations and combinations of the visual features will then be considered as possible construction moves in the next stage. This is achieved through a backtracking algorithm, where EuclidNet tries to build a solution by picking one possible construction move at a time as a new input or backtracking if it reaches the pre-defined maximum search depth. Since the goal construction is extracted during the segmentation process, our framework also examines new possible constructions that contain the targeted geometries. If a solution is found, the framework stops the procedure and saves the current search path. This entire procedure will repeat until there are no new possible constructions.}
    \label{fig:formalreasoning}
\end{figure}
\subsection{Backtracking Algorithm}
EuclidNet performs an exhaustive search on the space of all possible moves to enumerate all possible compositions of tools and verifies if any of the constructions match the given target construction. In an exhaustive search, we develop a backtracking algorithm to construct the geometries recursively to build the solution incrementally, one possible move at a time, and we remove those constructions that fail immediately. The final solution is derived from tracing backward and assembling the steps together to form an integrated whole. This backtracking procedure is described in Figure \ref{fig:formalreasoning}, and the detailed implementation of the algorithm can be found in Appendix \ref{appendixb}.

\subsection{Diagram Element Segmentation and Localization}
In our framework, we implement our EuclidNet for the segmentation and localization of primitives and intersections on top of Mask R-CNN with various backbones \cite{he2017mask}, i.e., Resnet-50 and Resnet-101. First, Mask R-CNN utilizes a regional proposal network to generate regional proposals of the images and determine whether the anchors are foreground or background. Next, the feature map and generated region proposals are forwarded to RoI align to generate a fixed-size feature map. Finally, three tasks are trained simultaneously: classification, box regression, and mask regression, which is performed to increase the accuracy of the mask. The total training loss can be defined as:
\begin{align}
    \mathcal{L} = \mathcal{L}_{cls} + \mathcal{L}_{box} + \mathcal{L}_{mask},
\end{align}
where $L_{cls}$ is the loss of the classification, $L_{box}$ is the bounding-box loss that is identified as those defined in \cite{girshick2015fast}, and $L_{mask}$ is the average binary cross-entropy loss including solely the $k$-th mask if the region is associated with the ground truth class $k$ from \cite{he2017mask}. There is a primitives segmentation model for extracting geometric primitives from the target and existing construction images respectively. Since it is still rather difficult to derive insights directly from the result when we construct a new move onto the existing constructions, we also implement an intersection segmentation model to emphasize the intersections of different geometries and treat them as extra points to be concatenated into the point segments, which helps the overall visual reasoning process.

\begin{figure}
  \centering
  \includegraphics[width=0.78\linewidth]{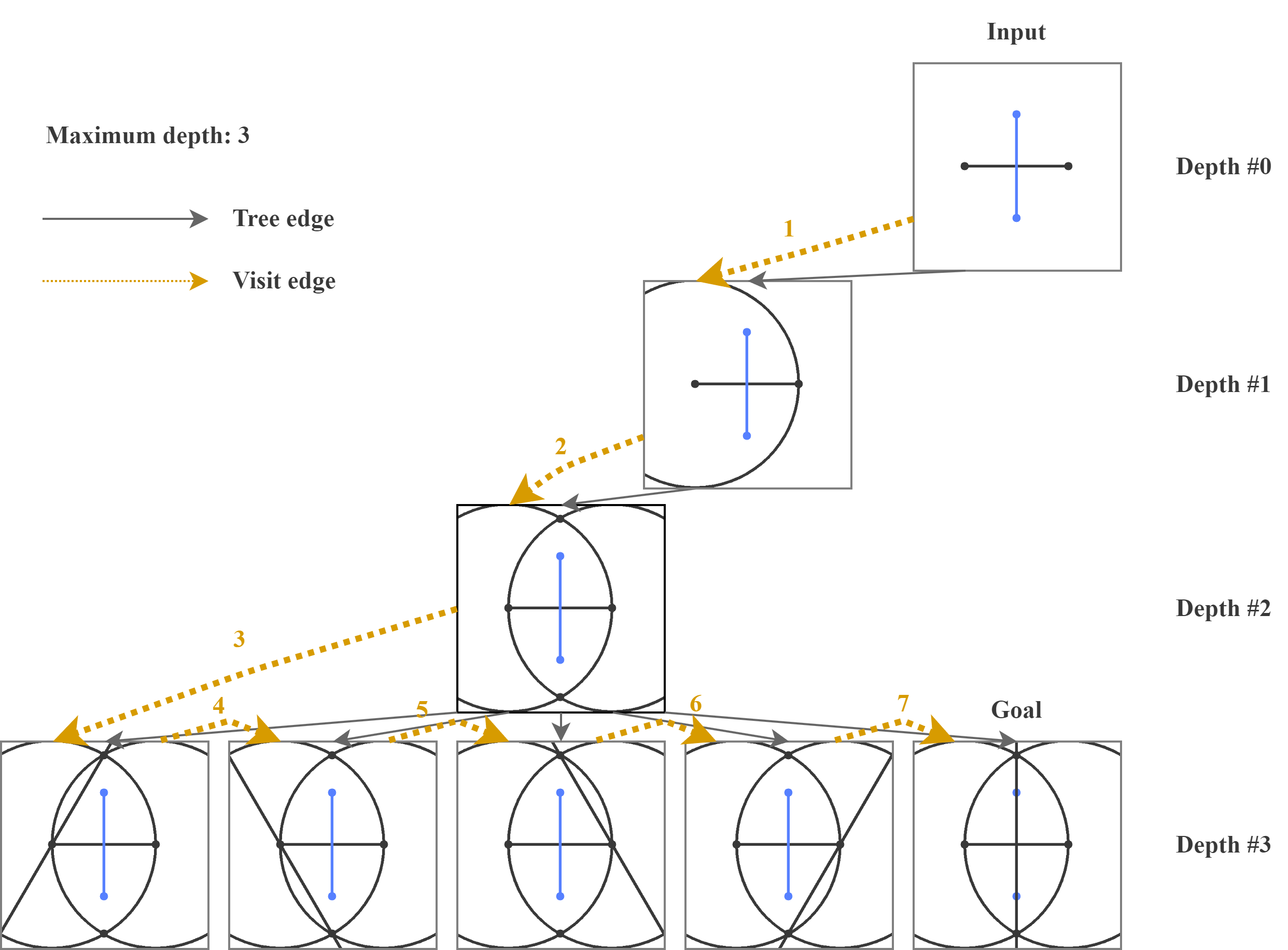}
  \caption{Illustration of the construction of a perpendicular bisector on a line. Given a line with two points as the existing configuration (in black), this problem prompts us to construct a perpendicular bisector (in blue) using only straightedges and compasses. When the search level reaches the maximum depth, the search will backtrack recursively with another tree traversal routine.}
  \label{fig-perpendicularbisector}
\end{figure}

\section{Experiment}
We build the dataset for primitives and intersections extraction by injecting different primitives and intersections with numbers and scales on images for various experiments with details given in Appendix \ref{appendixc1}. As a performance metric for geometric pattern perception, we adopt the mean average precision (mAP) as our evaluation metric to determine the effectiveness of our segmentation models. Detailed results are presented in Appendix \ref{appendixc2}. We also test EuclidNet with questions in Euclidea \cite{euclidea} to verify its effectiveness in solving geometric construction problems, and we see that it can accurately generate the solutions. Particularly, we also examine how effective EuclidNet is in recognizing patterns like isometries in its construction stages. One instance of the problems solved by EuclidNet is demonstrated in Figure \ref{fig-perpendicularbisector}. Moreover, EuclidNet demonstrates effectiveness in solving complex geometric construction questions such as the Japanese Sangaku problems \cite{pedoesangaku}. More problems solved by EuclidNet, including two Sangaku problems, can be found in Appendix \ref{appendixd}.

\section{Conclusion}
In this paper, we present EuclidNet, a novel deep learning-driven framework for visual reasoning on geometric constructible problems. The framework localizes geometric primitives using Mask R-CNN models and identifies new intersections from previous moves to discover new constructions. To find the solution, EuclidNet employs a backtracking algorithm to search for possible constructions with relatively low computational complexity. To validate EuclidNet's effectiveness to explore the solution space of constructible problems and its efficacy at reasoning `constructions', we have conducted various experiments on numerous challenging constructible problems, including Japanese Sangaku geometry. As future work, it is interesting to extend EuclidNet with neural-symbolic artificial intelligence or transfer learning that trains with a minimal amount of data to address geometric construction problems arranged in increasing order of difficulty.

\section*{Acknowledgment}
The work is supported in part by the Ministry of Education, Singapore, under its Academic Research Fund Tier 1 (Project No. 022307) and Hong Kong ITF Project No. ITS/188/20.

\medskip
\bibliography{neurips_2022.bib}

\begin{thebibliography}{23}
\providecommand{\natexlab}[1]{#1}
\providecommand{\url}[1]{\texttt{#1}}
\expandafter\ifx\csname urlstyle\endcsname\relax
  \providecommand{\doi}[1]{doi: #1}\else
  \providecommand{\doi}{doi: \begingroup \urlstyle{rm}\Url}\fi

\bibitem[Hilbert(1902)]{hilbert1902foundations}
David Hilbert.
\newblock \emph{The Foundations of Geometry}.
\newblock Open Court Publishing Company, 1902.

\bibitem[Pedoe(1988)]{pedoe}
Daniel Pedoe.
\newblock \emph{Geometry, a Comprehensive Course}.
\newblock Dover Books on Mathematics, 1988.

\bibitem[Gelernter and Rochester(1958)]{gelernter1958intelligent}
Herbert~L Gelernter and Nathaniel Rochester.
\newblock Intelligent behavior in problem-solving machines.
\newblock \emph{IBM Journal of Research and Development}, 2\penalty0
  (4):\penalty0 336--345, 1958.

\bibitem[Dreyfus(1991)]{dreyfus1991status}
Tommy Dreyfus.
\newblock On the status of visual reasoning in mathematics and mathematics
  education.
\newblock In \emph{Proc. 15th Conf. of the Int. Group for the Psychology of
  Mathematics Education}, 1991.

\bibitem[Kim(1989)]{kim1989}
Michelle~Y. Kim.
\newblock Visual reasoning in geometry theorem proving.
\newblock \emph{Proceedings of the 11th International Joint Conference on
  Artificial Intelligence}, 1989.

\bibitem[Koedinger and Anderson(1990)]{koedinger1990abstract}
Kenneth~R Koedinger and John~R Anderson.
\newblock Abstract planning and perceptual chunks: Elements of expertise in
  geometry.
\newblock \emph{Cognitive Science}, 14\penalty0 (4):\penalty0 511--550, 1990.

\bibitem[Gulwani et~al.(2011)Gulwani, Korthikanti, and
  Tiwari]{gulwani2011synthesizing}
Sumit Gulwani, Vijay~Anand Korthikanti, and Ashish Tiwari.
\newblock Synthesizing geometry constructions.
\newblock \emph{PLDI on ACM SIGPLAN}, 46\penalty0 (6):\penalty0 50--61, 2011.

\bibitem[Seo et~al.(2015)Seo, Hajishirzi, Farhadi, Etzioni, and
  Malcolm]{seo2015solving}
Minjoon Seo, Hannaneh Hajishirzi, Ali Farhadi, Oren Etzioni, and Clint Malcolm.
\newblock Solving geometry problems: Combining text and diagram interpretation.
\newblock In \emph{Proceedings of the 2015 Conference on Empirical Methods in
  Natural Language Processing}, pages 1466--1476, 2015.

\bibitem[Bronstein et~al.(2021)Bronstein, Bruna, Cohen, and
  Velickovic]{geometricdeeplearning}
M.~M. Bronstein, J.~Bruna, T.~Cohen, and P.~Velickovic.
\newblock Geometric deep learning: Grids, groups, graphs, geodesics, and
  gauges.
\newblock \url{https://arxiv.org/abs/2104.13478}, 2021.

\bibitem[Davies et~al.(2021)Davies, Velickovic, Buesing, and
  et~al.]{naturemathai}
A.~Davies, P.~Velickovic, L~Buesing, and et~al.
\newblock Advancing mathematics by guiding human intuition with ai.
\newblock \emph{Nature}, 600:\penalty0 70--74, 2021.

\bibitem[Wang(1984)]{wang1990computer}
Hao Wang.
\newblock Computer theorem proving and artificial intelligence.
\newblock In \emph{Automated Theorem Proving: After 25 Years}, volume~29, pages
  49--–70. Springer, 1984.

\bibitem[Wang(1960{\natexlab{a}})]{wangibm}
Hao Wang.
\newblock Toward mechanical mathematics.
\newblock In \emph{IBM Journal}, volume~4, pages 2--–22, 1960{\natexlab{a}}.

\bibitem[Wang(1960{\natexlab{b}})]{wangacm}
Hao Wang.
\newblock Proving theorems by pattern recognition.
\newblock In \emph{Part I, Communications of the Association for Computing
  Machinery}, volume~3, pages 220--–234, 1960{\natexlab{b}}.

\bibitem[Fukagawa and Pedoe(1989)]{pedoesangaku}
Hidetoshi Fukagawa and Daniel Pedoe.
\newblock \emph{Japanese Temple Geometry Problems}.
\newblock Charles Babbage Research Centre, 1989.
\newblock ISBN 9780919611214.

\bibitem[Geretschl{\"a}ger(1995)]{geretschlager1995euclidean}
Robert Geretschl{\"a}ger.
\newblock Euclidean constructions and the geometry of origami.
\newblock \emph{Mathematics Magazine}, 68\penalty0 (5):\penalty0 357--371,
  1995.

\bibitem[LeCun et~al.(2015)LeCun, Bengio, and Hinton]{lecun2015deep}
Yann LeCun, Yoshua Bengio, and Geoffrey Hinton.
\newblock Deep learning.
\newblock \emph{Nature}, 521\penalty0 (7553):\penalty0 436--444, 2015.

\bibitem[He et~al.(2017)He, Gkioxari, Doll{\'a}r, and Girshick]{he2017mask}
Kaiming He, Georgia Gkioxari, Piotr Doll{\'a}r, and Ross Girshick.
\newblock Mask {R-CNN}.
\newblock In \emph{Proceedings of the IEEE International Conference on Computer
  Vision}, pages 2961--2969, 2017.

\bibitem[Girshick(2015)]{girshick2015fast}
Ross Girshick.
\newblock Fast {R-CNN}.
\newblock In \emph{Proceedings of the IEEE international conference on computer
  vision}, pages 1440--1448, 2015.

\bibitem[euc()]{euclidea}
Euclidea.
\newblock URL \url{https://www.euclidea.xyz}.

\bibitem[Lu et~al.(2021)Lu, Gong, Jiang, Qiu, Huang, Liang, and
  Zhu]{lu2021inter}
Pan Lu, Ran Gong, Shibiao Jiang, Liang Qiu, Siyuan Huang, Xiaodan Liang, and
  Song-Chun Zhu.
\newblock Inter-{GPS}: Interpretable geometry problem solving with formal
  language and symbolic reasoning.
\newblock In \emph{The 59th Annual Meeting of the Association for Computational
  Linguistics}, 2021.

\bibitem[Gao and Chou(1998)]{gao1998solving}
Xiao-Shan Gao and Shang-Ching Chou.
\newblock Solving geometric constraint systems. {II}. a symbolic approach and
  decision of {RC}-constructibility.
\newblock \emph{Computer-aided design}, 30\penalty0 (2):\penalty0 115--122,
  1998.

\bibitem[Macke et~al.(2021)Macke, Sedlar, Olsak, Urban, and
  Sivic]{macke2021learning}
Jaroslav Macke, Jiri Sedlar, Miroslav Olsak, Josef Urban, and Josef Sivic.
\newblock Learning to solve geometric construction problems from images.
\newblock In \emph{International Conference on Intelligent Computer
  Mathematics}, pages 167--184. Springer, 2021.

\bibitem[Fukagawa and Rothman(2008)]{sangaku}
Hidetoshi Fukagawa and Tony Rothman.
\newblock \emph{Sacred Mathematics: Japanese Temple Geometry}.
\newblock Princeton University Press, 2008.
\newblock ISBN 9780691127453.

\end{thebibliography}

\section{Appendix}

\subsection{Related Works}
\subsubsection{Computational Geometry}
\paragraph{Automated Geometric Reasoning with Formal Reasoning} 
Automated geometric reasoning refers to the process of deriving theorems from geometric problems with computers, which was first attempted by Gelernter and his collaborators in 1959 \cite{gelernter1958intelligent}. Existing automated geometric reasoning methods primarily utilize formal reasoning, which relies on rules of logic and mathematics to conduct reasoning. These methods can be further categorized into synthetic and algebraic approaches. For synthetic reasoning, geometric definitions and theorems are built inside machines, which then incorporate search techniques, often exhaustive, to find the solution \cite{gulwani2011synthesizing}. Examples of the synthetic approach can be found in Inter-GPS \cite{lu2021inter} and GEOS \cite{seo2015solving}, which are solvers with built-in Euclidean formulas. 
However, the algebraic approach has many limitations, and no geometric construction solver has adopted this method thus far. 

\paragraph{Automated Geometric Reasoning on Constructions} An early attempt to perform geometric reasoning on constructions with images was made by Gao \cite{gao1998solving} to extract information from the diagram and produce a construction sequence. A more recent image-based geometric construction problem solver is built on top of Mask R-CNN for finding geometric constructions with straightedges and compasses in the Euclidea \cite{macke2021learning}. The solver uses a Mask R-CNN as a recognizer to generate geometric meanings and arrive at the solutions without using formal reasoning. To be more specific, their approach trains the deep learning models by the images from different tools on Euclidea with variations such as rotation and translation as the dataset. However, this image-based solver only performs effectively on the training data from Euclidea.

\subsection{Geometric Construction with Straightedges and Compasses} \label{appendixa2}
\paragraph{Geometric Construction Environment on Euclidea} With the burgeoning development of online education platforms, gamified applications like Euclidea \cite{euclidea} are more widely adopted as they have relatively large collections of problems in the game databases. Our geometric construction environment follows a similar setting with only point, line, and circle tools. Some advanced tools available on Euclidea as shortcuts for complex constructions, such as perpendicular bisector, angle bisector, perpendicular, and parallel, can also be constructed using basic tools (see Figure \ref{fig-perpendicularbisector}). Users are asked to find a sequence of construction steps from an existing configuration (in black) to a given goal configuration (in blue) using the rule and compass. 


\paragraph{Points of Intersections} Since moves in our configuration rely on the existing points in the Euclidean plane, it is required to define new possible points in the plane, i.e., intersections and points from the ray line from the previous moves. For the two non-parallels, there is not necessarily an intersection point for a line segment, but an \textit{line-line intersection} point must exist with their ray. Secondly, we consider the \textit{line-circle intersection}. There are three ways a line and a circle can be associated, i.e., the line cuts the circle at two distinct points, the line is tangent to the circle, or the line misses the circle. Finally, we define the cases on \textit{circle-circle intersection}. If the sum of the radii and the distance between the centers are equal, then the circles touch externally. Also, if the difference between the radii and the distance between the centers are equal, then the circles touch internally. The summary for points of intersections is as follows:

\begin{enumerate}
    \item Given two non-parallel straight lines $l_1$ and $l_2$, we can determine their unique point of intersection $P=l_1 \cap l_2$.
    
    \item Given a circle $c =\{C;r\}$ and a straight line $l$, the distance between $C$ and $l$ less than or equal to $r$, there will be either two points or one point of intersection between $c$ and $l$.
    
    \item Given two circles $c_1=\{C_1;r_1\}$ and $c_2=\{C_2;r_2\}$ such that neither contains the center of the other in its interior, and the distance between the centers is less than or equal to the sum of the radii, then there will be two points of intersection between $c_1$ and $c_2$.
    
    \item Given two circles $c_1=\{C_1;r_1\}$ and $c_2=\{C_2;r_2\}$ such that one contains the center of the other in its interior, and the distance between the centers is no less than the difference between the two radii, there will be at most four points of intersection between $c_1$ and $c_2$.
\end{enumerate}

\paragraph{Construction on Euclidea Puzzle: Inscribed Square}
Figure 3 illustrates a solution to a problem from Euclidea (Alpha-7) with the initial configuration in the black channel and the goal configuration in the blue channel. Users are asked to draw the inscribed square of a circle given the circle with its center. In the original setting of Euclidea, we can solve this problem in 6 steps with shortcut tools such as the perpendicular bisector. However, we show an optimal solution for this problem with only line and circle tools. At each step, the black channel is updated and registered as the current state of the problem. This process will repeat until the current state is identical to the goal configuration.
\begin{figure}[h]
\begin{minipage}[t]{0.24\linewidth}
\centering
    \includegraphics[width=\linewidth]{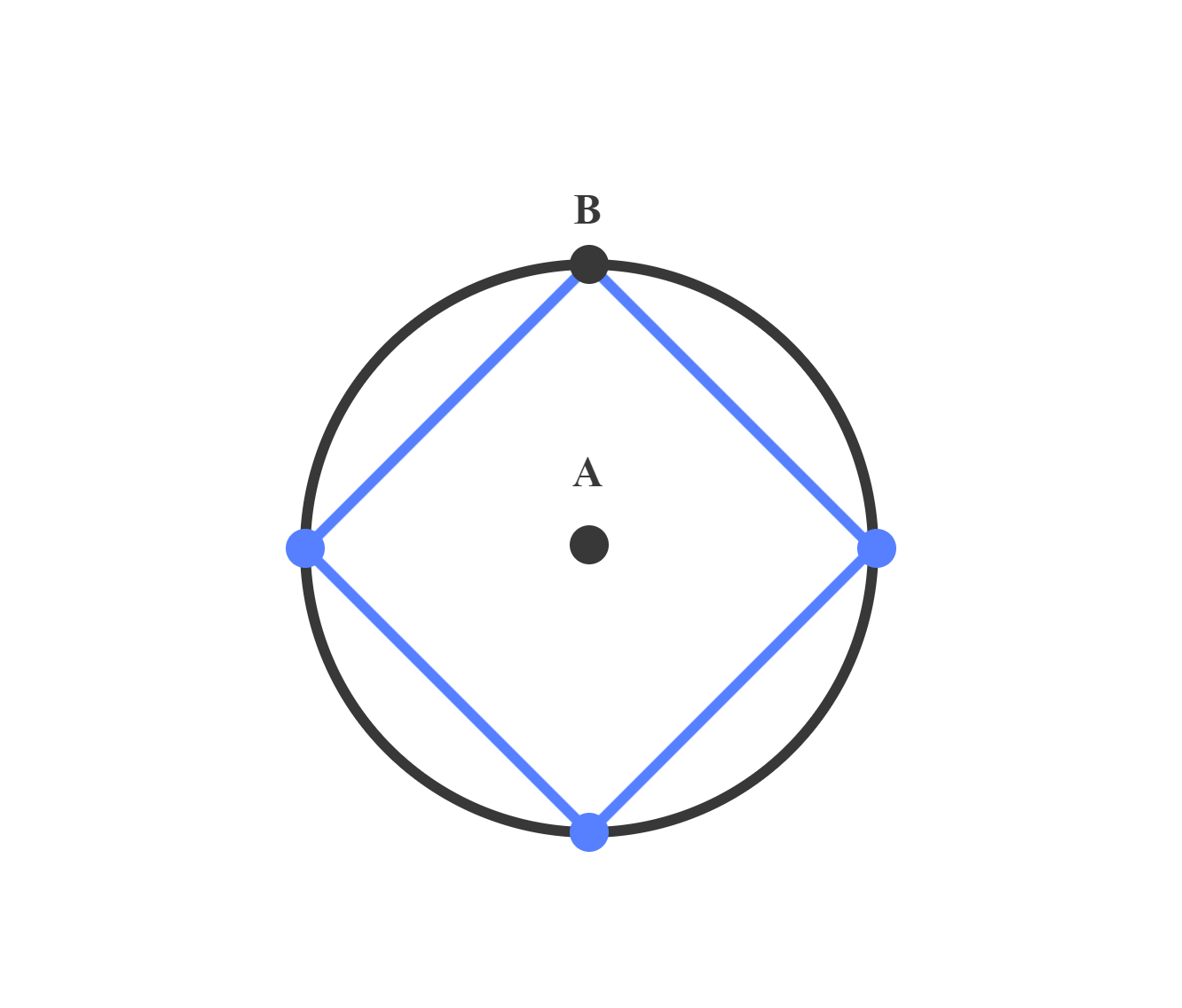}%
\label{fig_first_case}
\subcaption*{\textbf{Input}: Euclidea \textit{Alpha}-7 (Inscribed Square): Given a circle with center $A$ and a point $B$ on circle, inscribe a square in the circle.}
\end{minipage}
\hspace{0.1em}
\vspace{0.1em}
\begin{minipage}[t]{0.24\linewidth}
    \centering
    \includegraphics[width=\linewidth]{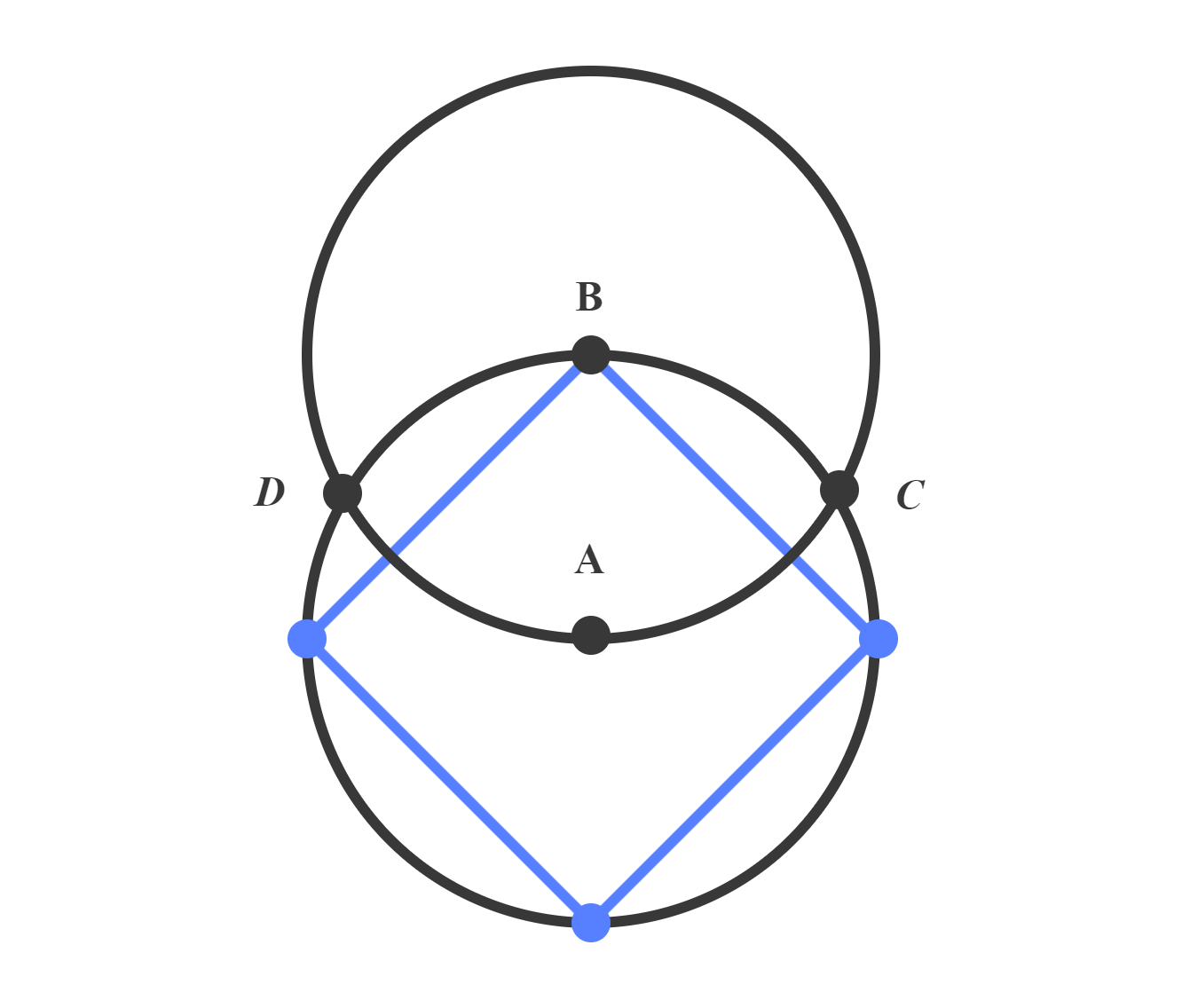}%
    \label{fig_second_case}
\subcaption*{\textbf{Step 1}: Draw a circle from $A$ as a center to $B$. Two points $C$ and $D$ from circle-circle intersection will be obtained.}
\end{minipage}
\hspace{0.1em}
\vspace{0.1em}
\begin{minipage}[t]{0.24\linewidth}
    \centering
    \includegraphics[width=\linewidth]{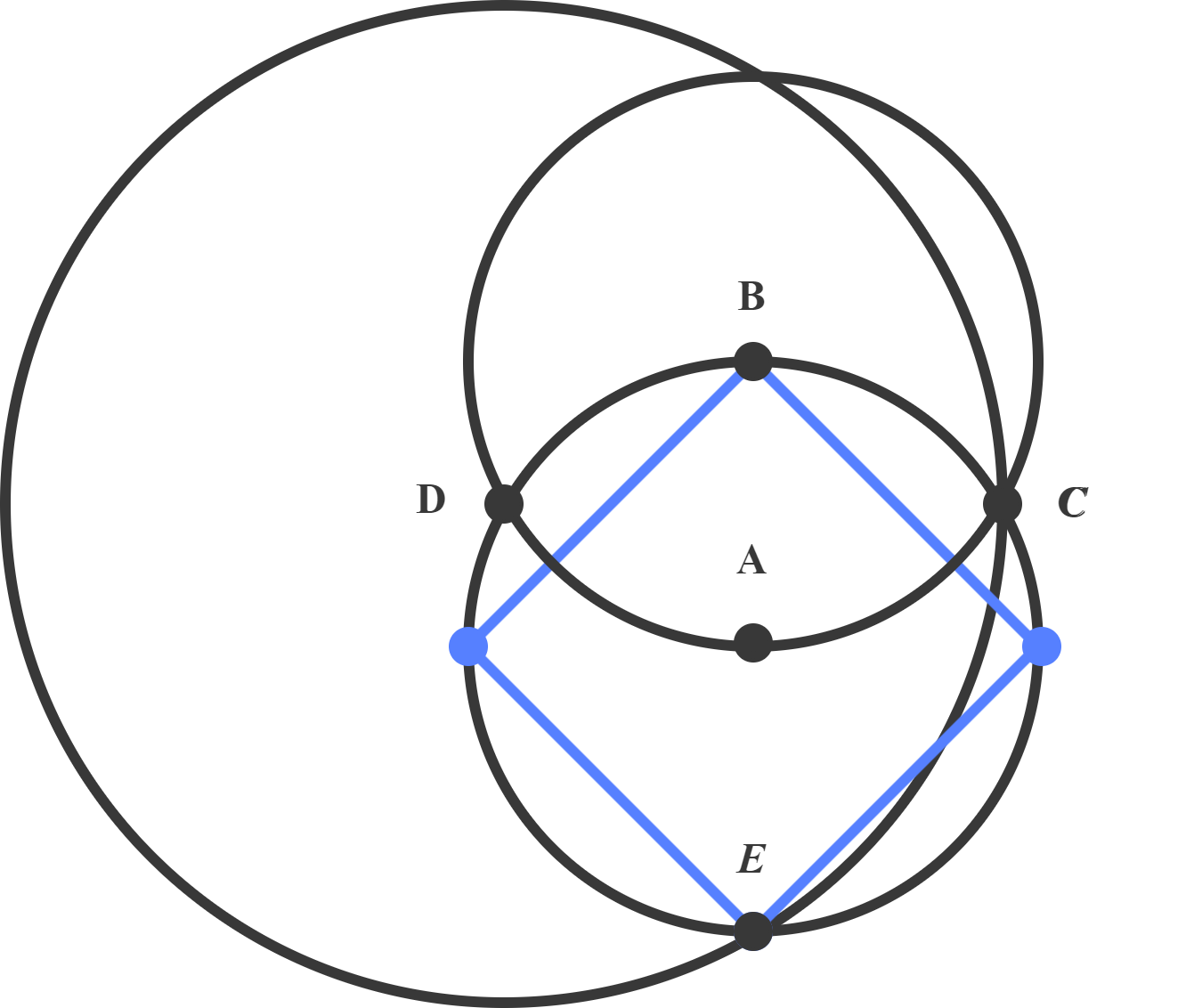}%
    \label{fig_third_case}
\subcaption*{\textbf{Step 2}: Draw a circle from $D$ as a center to $C$. A point $E$ from circle-circle intersection will be obtained.}
\end{minipage} 
\hspace{0.1em}
\vspace{0.1em}
\begin{minipage}[t]{0.24\linewidth}
    \centering
    \includegraphics[width=\linewidth]{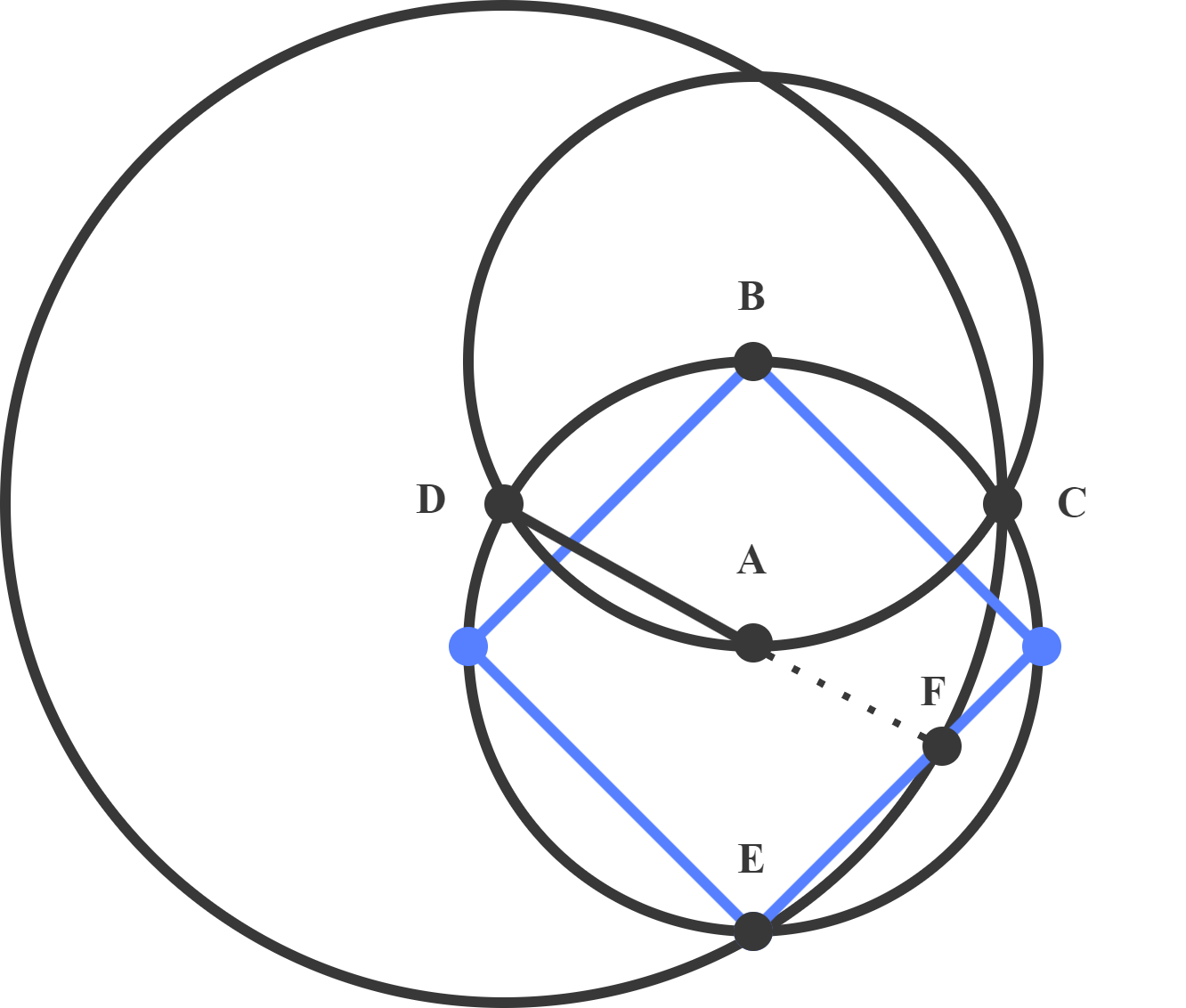}%
    \label{fig_fourth_case}
\subcaption*{\textbf{Step 3}: Draw a line from $D$ to $A$, a point $F$ will be obtained from a ray of $DA$. }
\end{minipage} 
\hspace{0.1em}
\vspace{0.1em}
\begin{minipage}[t]{0.24\linewidth}
    \centering
    \includegraphics[width=\linewidth]{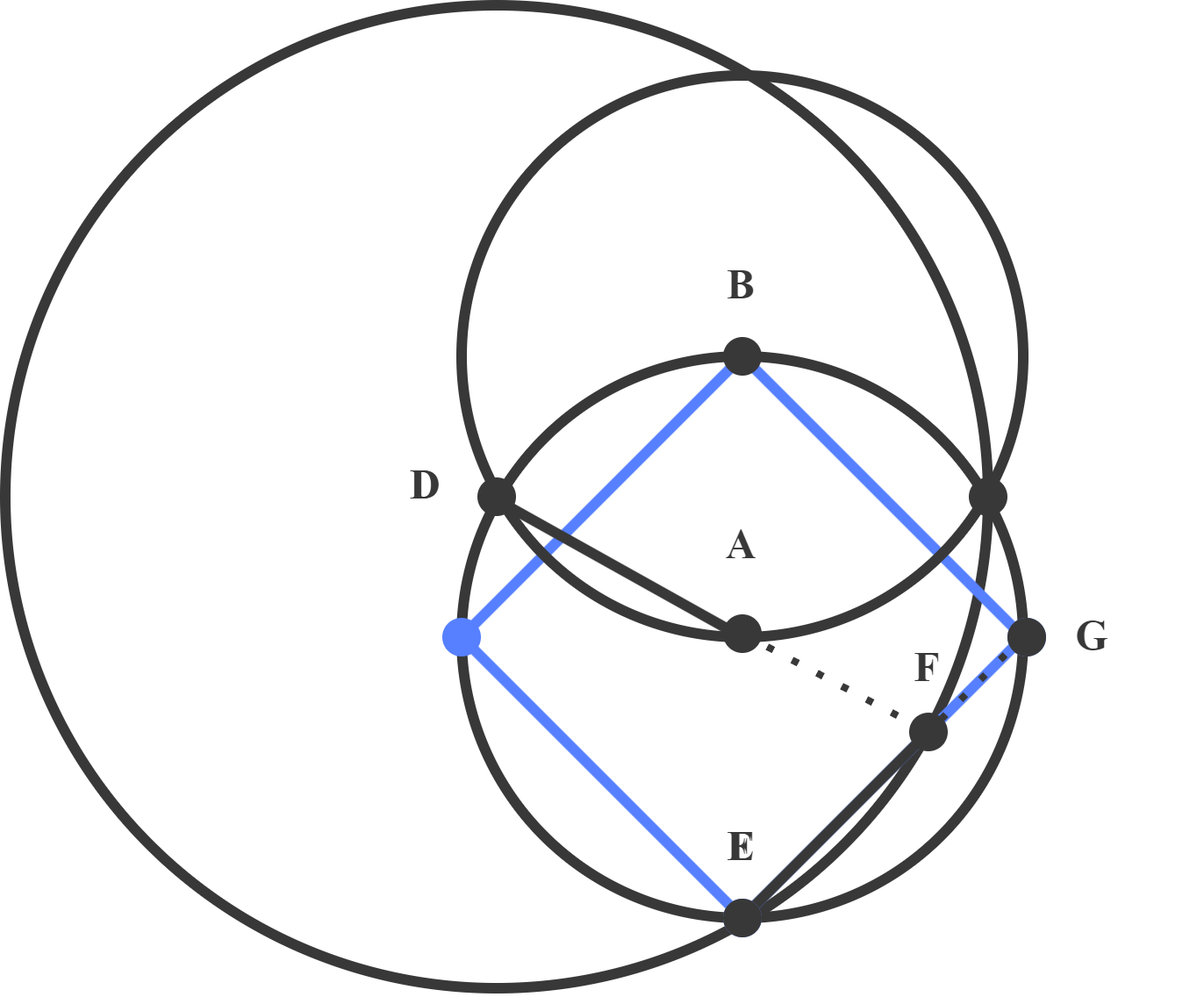}%
    \label{fig_fifth_case}
\subcaption*{\textbf{Step 4}: Draw a line from $E$ to $F$, a point $G$ will be obtained from a ray of $EF$.}
\end{minipage} 
\hspace{0.1em}
\vspace{0.1em}
\begin{minipage}[t]{0.24\linewidth}
    \centering
    \includegraphics[width=\linewidth]{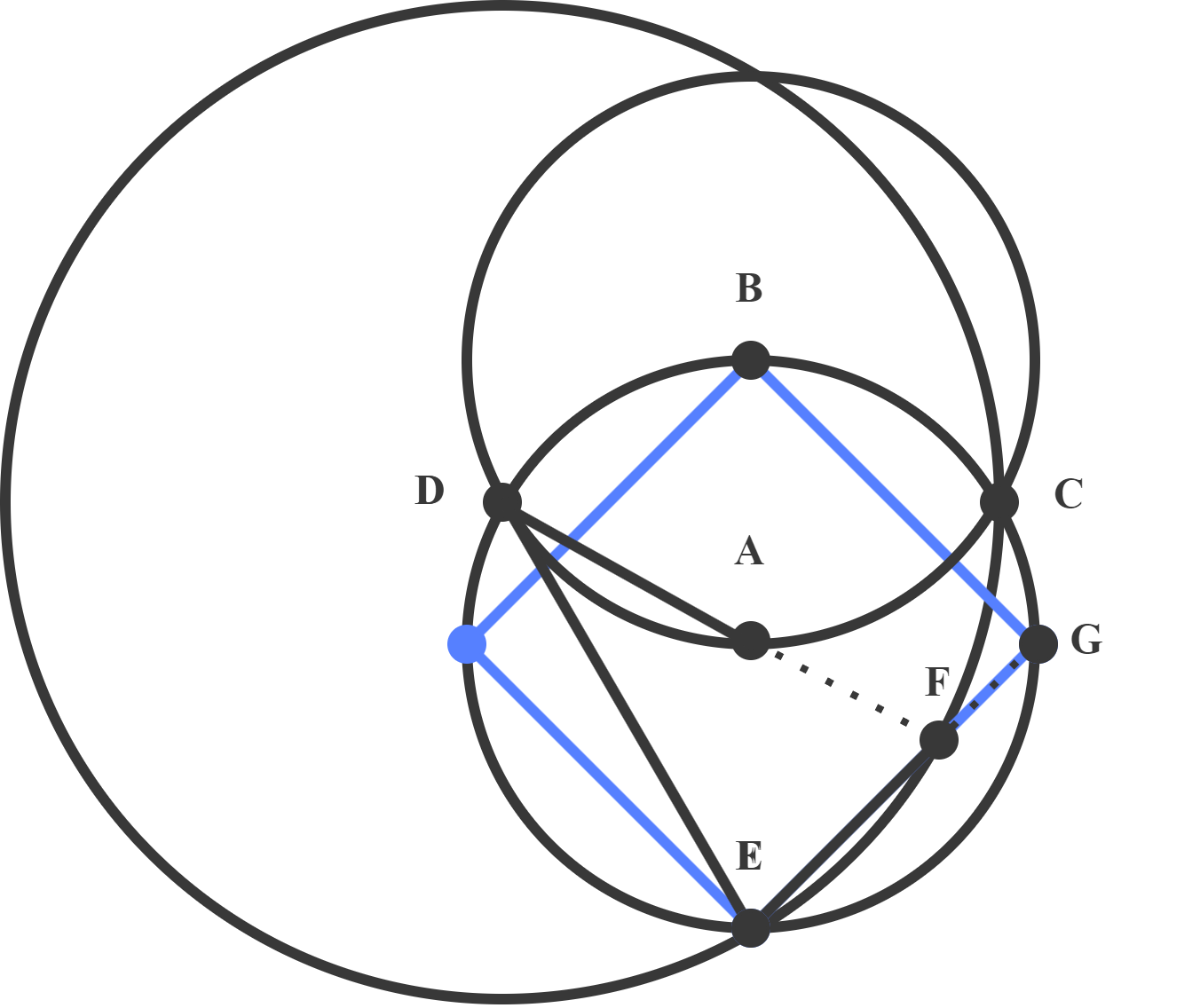}%
    \label{fig_sixth_case}
\subcaption*{\textbf{Step 5}: Draw a line from $D$ to $E$.}
\end{minipage} 
\hspace{0.1em}
\vspace{0.1em}
\begin{minipage}[t]{0.24\linewidth}
    \centering
    \includegraphics[width=\linewidth]{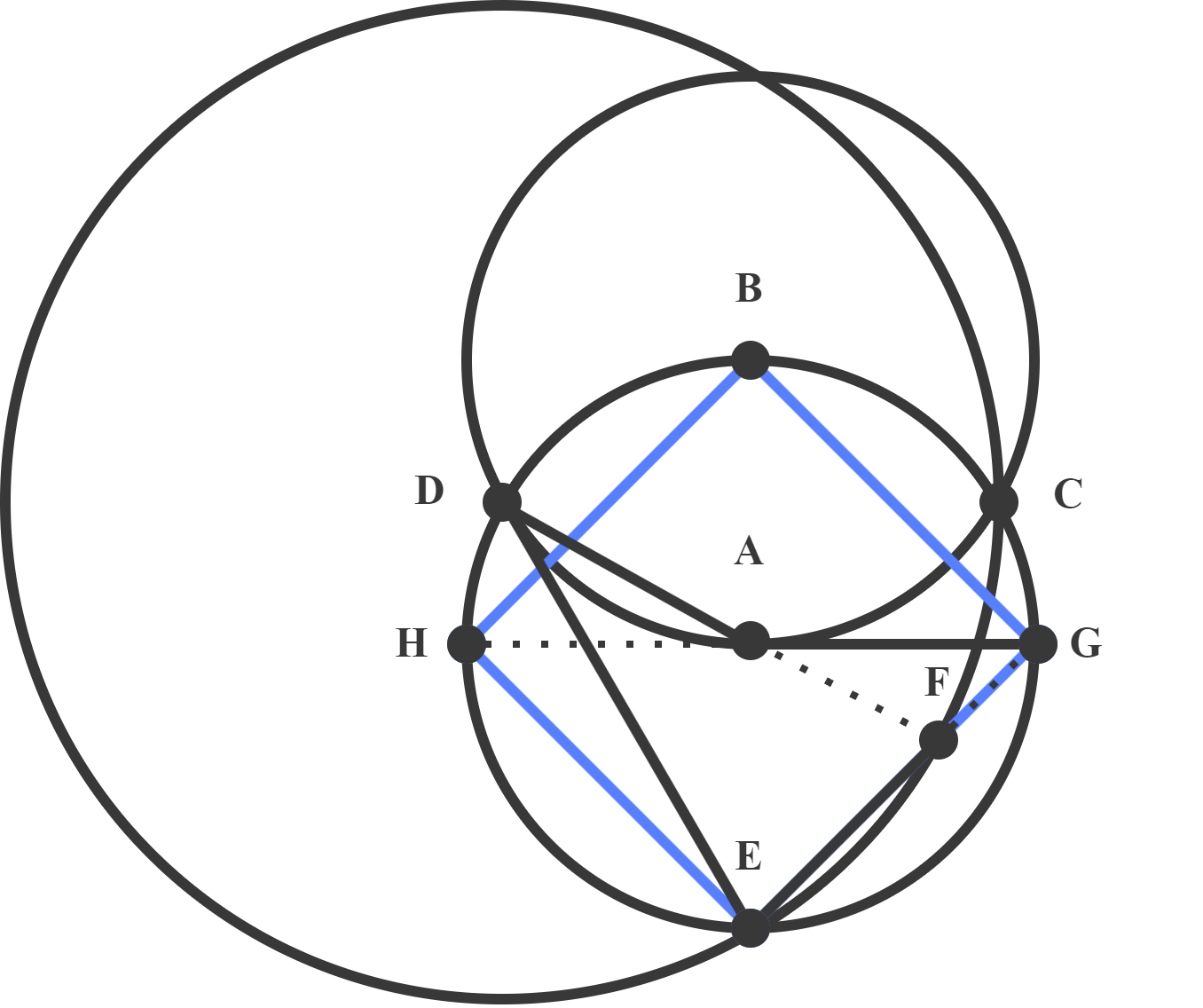}%
    \label{fig_seventh_case}
\subcaption*{\textbf{Step 6}: Draw a line from $G$ to $A$, a point $H$ will be obtained from a ray $AG$.}
\end{minipage} 
\hspace{0.1em}
\vspace{0.1em}
\begin{minipage}[t]{0.24\linewidth}
    \centering
    \includegraphics[width=\linewidth]{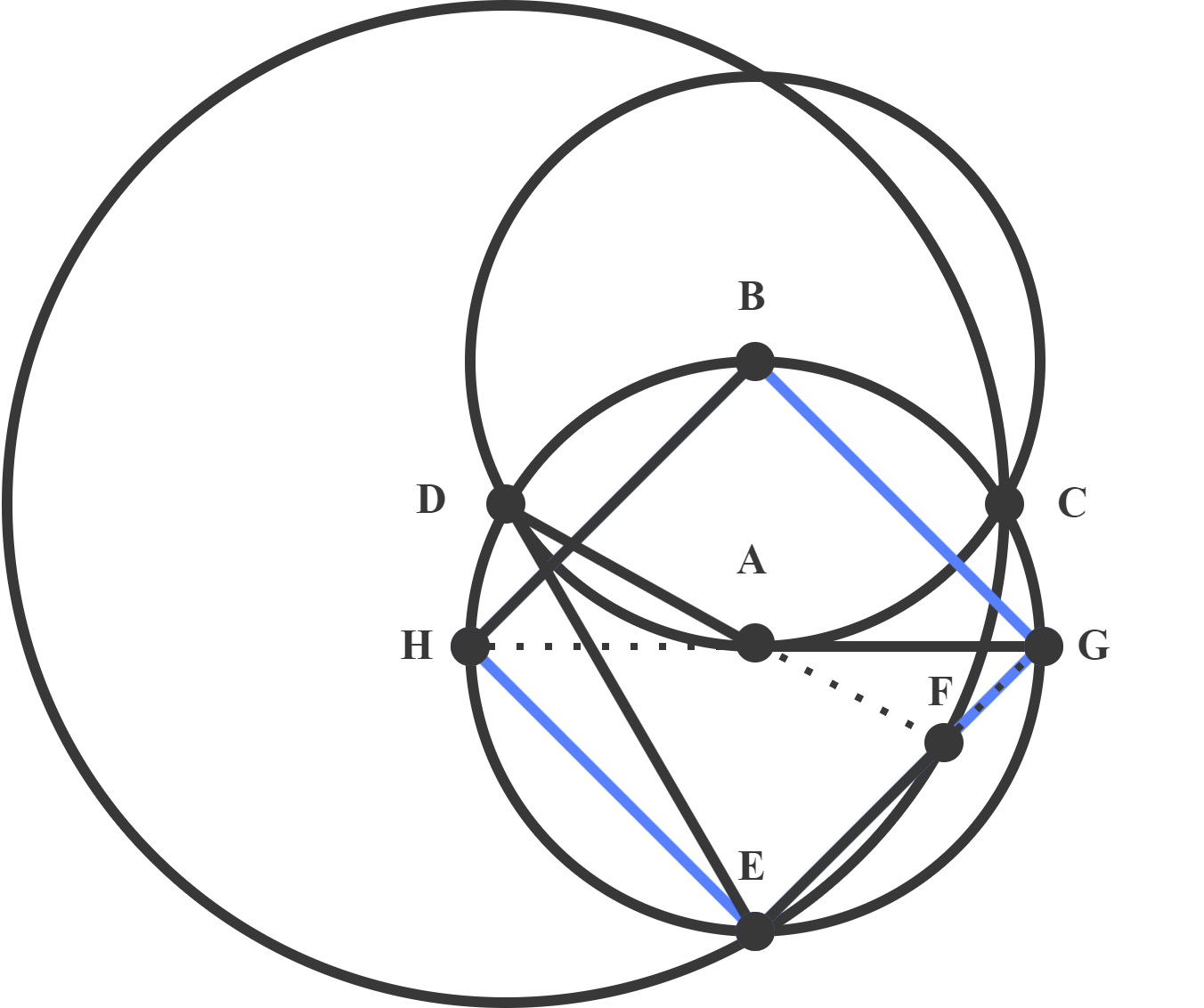}%
    \label{fig_eighth_case}
\subcaption*{\textbf{Step 7}: Draw a line from $H$ to $B$.}
\end{minipage} 
\hspace{0.1em}
\vspace{0.1em}
\begin{minipage}[t]{0.24\linewidth}
    \centering
    \includegraphics[width=\linewidth]{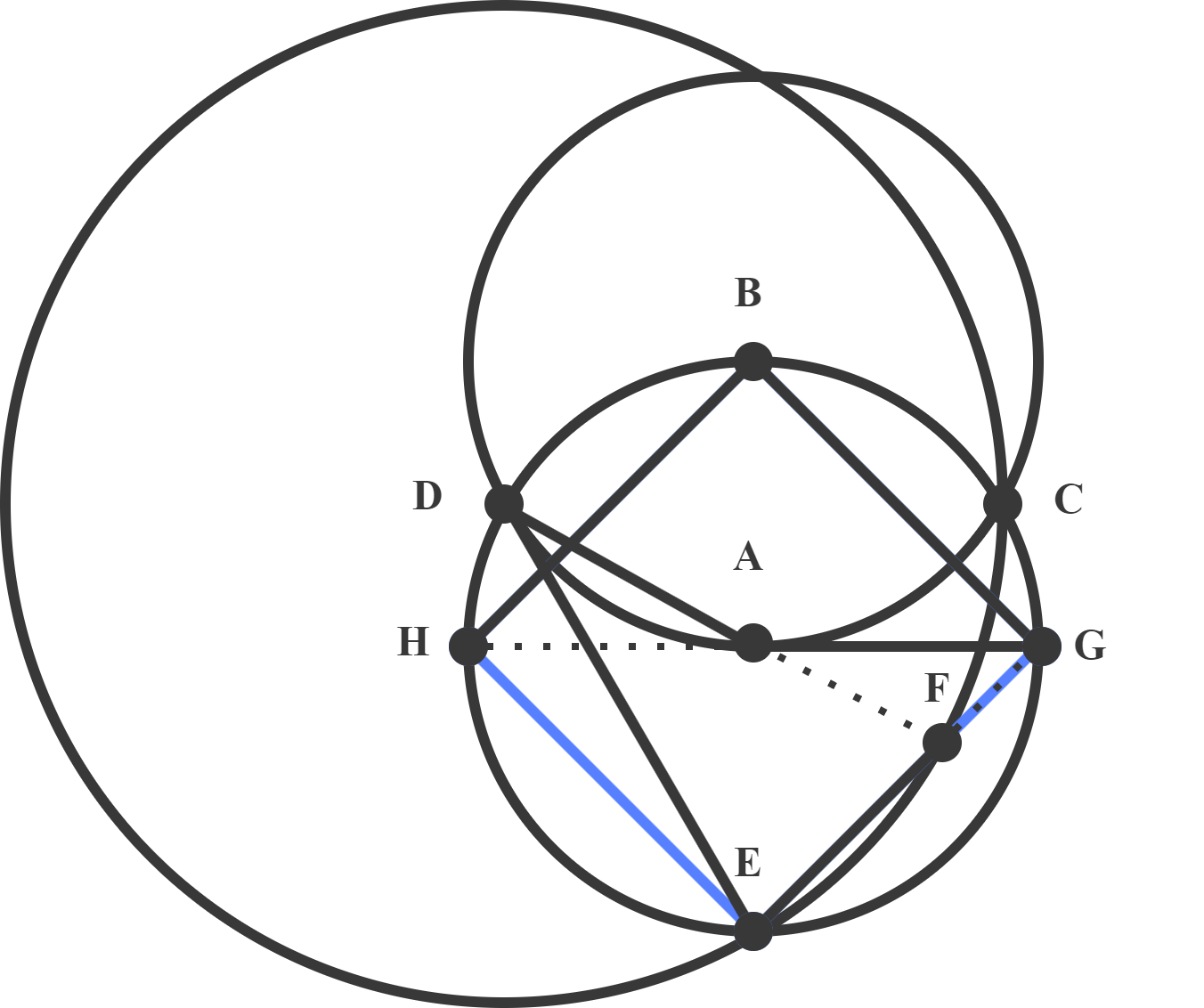}%
    \label{fig_fifth_case}
\subcaption*{\textbf{Step 8}: Draw a line from $B$ to $G$.}
\end{minipage} 
\hspace{0.1em}
\vspace{0.1em}
\begin{minipage}[t]{0.24\linewidth}
    \centering
    \includegraphics[width=\linewidth]{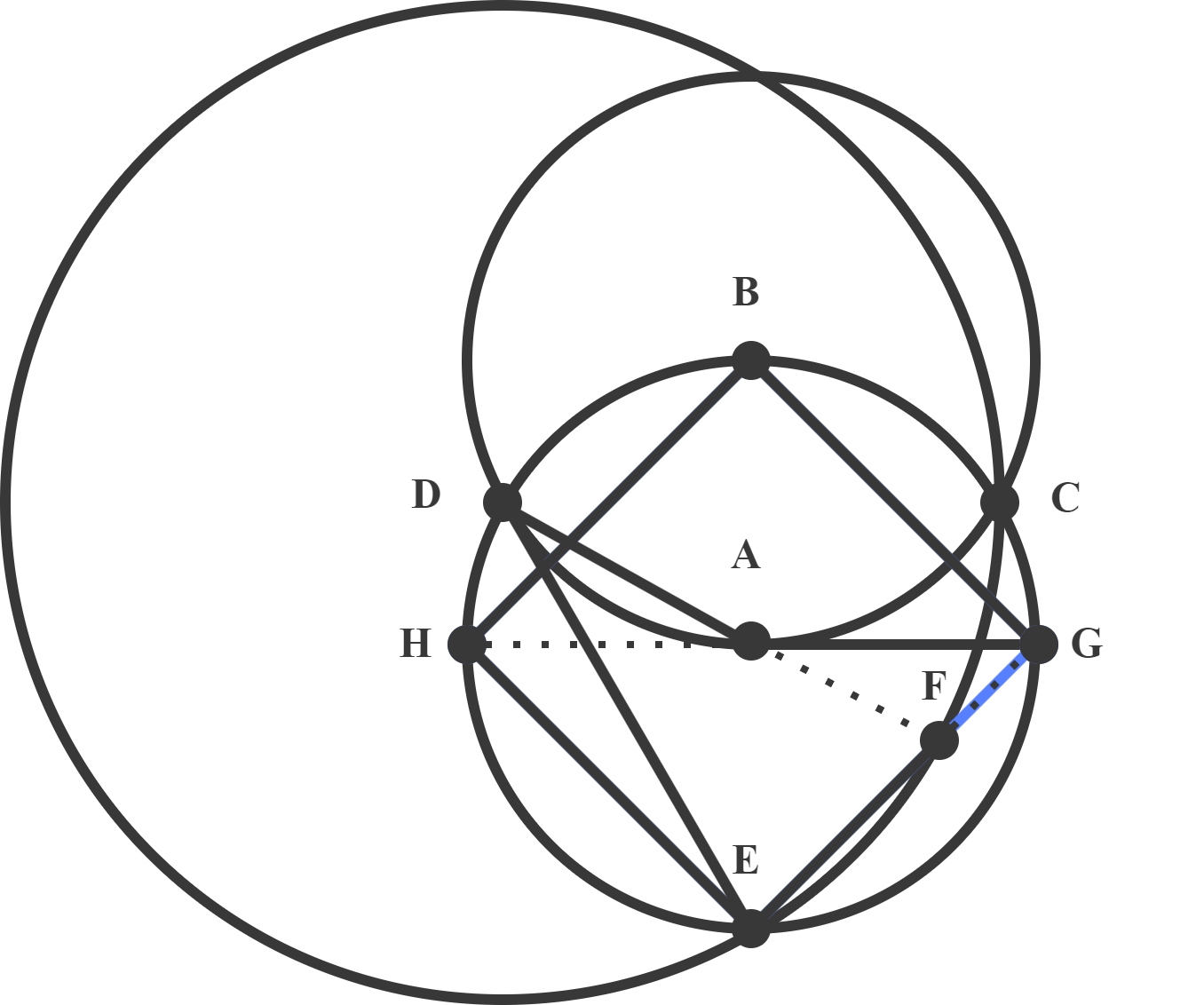}%
    \label{fig_fifth_case}
\subcaption*{\textbf{Step 9}: Draw a line from $H$ to $E$ and the construction is completed.}
\end{minipage} 
\hspace{0.1em}
\vspace{0.1em}
\caption{Illustration of construction on the inscribed square in a circle using straightedges and compasses. EuclidNet uses the same geometric construction environment described above. }
\end{figure}

\newpage
\section{Detailed Implementation} \label{appendixb}
\subsection{Deep Visual Reasoning with Backtracking}
The search algorithm takes an image of a constructible problem as the initial input. The primitives and intersections information will be extracted by the pre-trained segmentation models for each input image of the search simultaneously. Alternatively, the information can be carried forward to the next depth and skip the extraction during the search. The search will stop if a solution is found, and the sequence of images will then be saved.
\begin{algorithm}
\caption{(\textbf{DVRB}) \textbf{D}eep \textbf{V}isual \textbf{R}easoning with \textbf{B}acktracking }\label{alg:reasoning}
\hspace*{\algorithmicindent} \textbf{Input:} Image $I_{curr}$, Depth $D_{curr}$, Sequence of Images $\textit{\textbf{S}}$\\
\hspace*{\algorithmicindent} \textbf{Output:} Sequence of Images $\textit{\textbf{S}}$\\
\hspace*{\algorithmicindent} \textbf{Variables:}  Possible Moves $\textit{\textbf{M}}$, Points $\textit{\textbf{P}}$, Lines $\textit{\textbf{L}}$, Circles $\textit{\textbf{C}}$\\
\hspace*{\algorithmicindent} \textbf{Parameters:} Goal Image $I_{goal}$, Maximum Depth $D_{max}$\\
\hspace*{\algorithmicindent} \textbf{Models:} Primitives $Seg_{p}$ , Intersections $Seg_{i}$ \Comment{Pre-trained segmentation models}
\begin{algorithmic}
\If{ $I_{curr} = I_{goal}$} \Comment{Check $I_{curr}$ is solved}
    \State save($\textit{\textbf{S}}$)  \Comment{Save the solution}
    \State \textbf{return}
\Else
    \If{ $D_{curr} = D_{max}$} 
        \State \textbf{return} \Comment{Reach the maximum depth}
    \Else
        \State $\textit{\textbf{P}}$, $\textit{\textbf{L}}$, $\textit{\textbf{C}}$ $\leftarrow$ $Seg_{p}$.extract($I_{curr}$) \Comment{Extract primitives}
        \State $\textit{\textbf{P}}$ $\leftarrow$ $\textit{\textbf{P}}$ + $Seg_{i}$.extract($I_{curr}$) \Comment{Extract intersections for existing points}
        \State $\textit{\textbf{M}}$ $\leftarrow$ construct($\textit{\textbf{P}}$)  \Comment{Construct all moves defined in section 2.1}
        \State $\textit{\textbf{M}}$ $\leftarrow$ $\textit{\textbf{M}}$.remove($\textit{\textbf{L}},\textit{\textbf{C}}$) \Comment{Remove existing lines and circles from possible moves}
        \While{$\textit{\textbf{M}}$ not empty}
            \State $m\leftarrow M_0$ \Comment{Pick the first possible move in $\textit{\textbf{M}}$}
            \State $\textit{\textbf{S}}$ $\leftarrow I_{curr}\oplus m$ \Comment{Add the concatenation of a new move and current image}
            \State DVRB($I_{curr}\oplus m,D_{curr}+1,\textit{\textbf{S}}$) \Comment{Substitute new moves to next depth}
            \State $\textit{\textbf{S}}$.remove($I_{curr}\oplus m$) \Comment{Remove a tested construction}
            \State $\textit{\textbf{M}}$.remove($m$) \Comment{Remove a tested move}
        \EndWhile
        \State \textbf{return}
    \EndIf
\EndIf
\end{algorithmic}
\end{algorithm}
\subsection{Network Architecture}
We implement our EuclidNet for the segmentation and localization of primitives and intersections on top of Mask R-CNN \cite{he2017mask}. The input RGB image will be divided into two images for each channel and carried forward to the network separately. The detailed architecture is summarized as follows: 
\begin{figure}[!h]
  \centering
  \includegraphics[width=\linewidth]{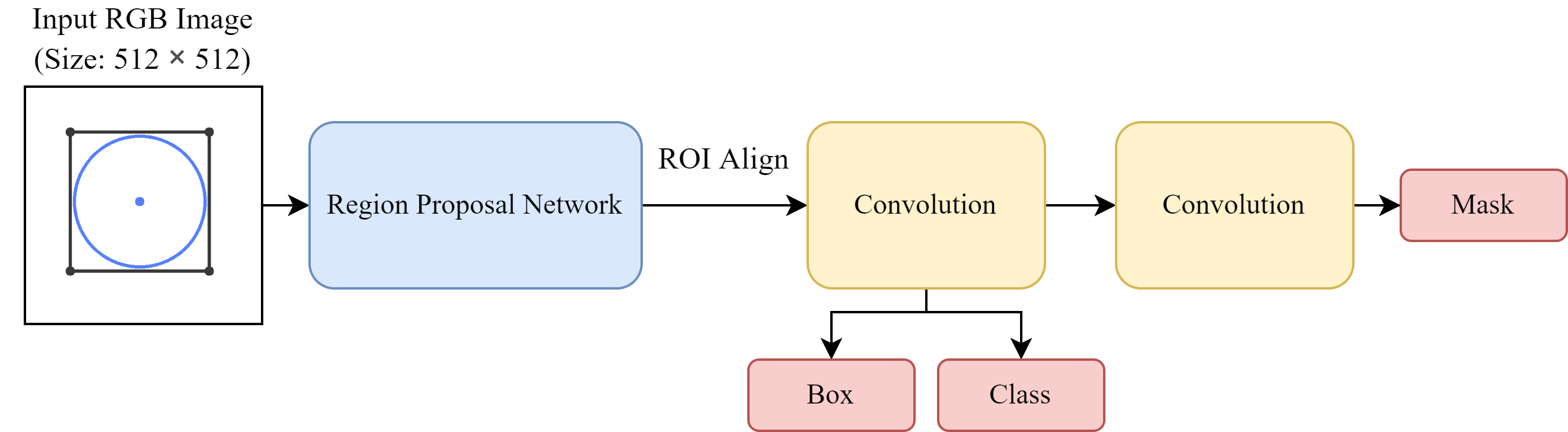}
  \caption{Illustration of Mask R-CNN architecutre in our framework.} 
\end{figure}
\newpage
\section{Experimental Details} 
\subsection{Experiment Setup} \label{appendixc1}
We evaluate the performance of our models in terms of the mean average precision (mAP) for both primitives and intersections. For the backbone network, we experiment with ResNet-50 and ResNet-101. To train segmentation models, we set the SGD optimizer with a learning rate of $0.001$ and the minimum detection confidence as $0.7$. The training is launched on a single NVIDIA 2080Ti GPU (11GB) with a batch size of 16. Other parameters follow the default configuration in \cite{he2017mask}. The models are trained on $1000$ simulating images in Figure 5 with primitives and intersections separately on $200$ epochs and $1000$ steps per epoch. We also evaluate the accuracy of EuclidNet on the first $6$ levels of Euclidea.
\begin{figure}[!h]
  \centering
  \includegraphics[width=0.7\linewidth]{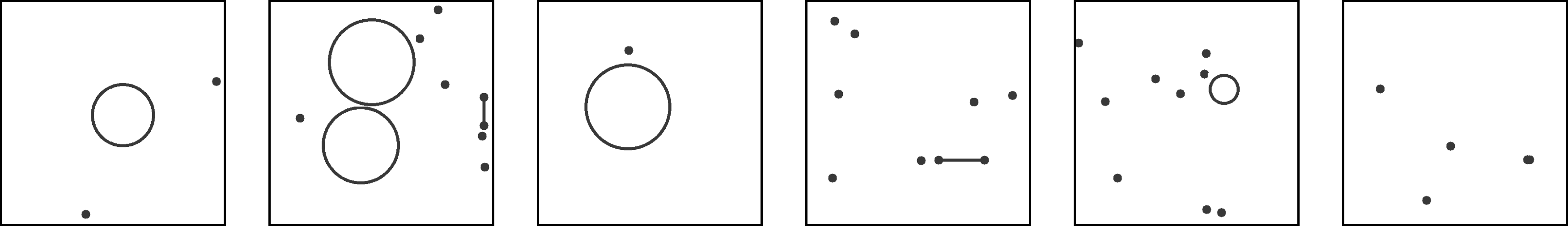}
  \caption{Illustration of injecting different primitives and intersections with numbers and scales.}
  \label{fig-sample}
\end{figure}

\subsection{Experiment Results} \label{appendixc2}
\begin{figure}[!h]
  \centering
  \includegraphics[width=0.78\linewidth]{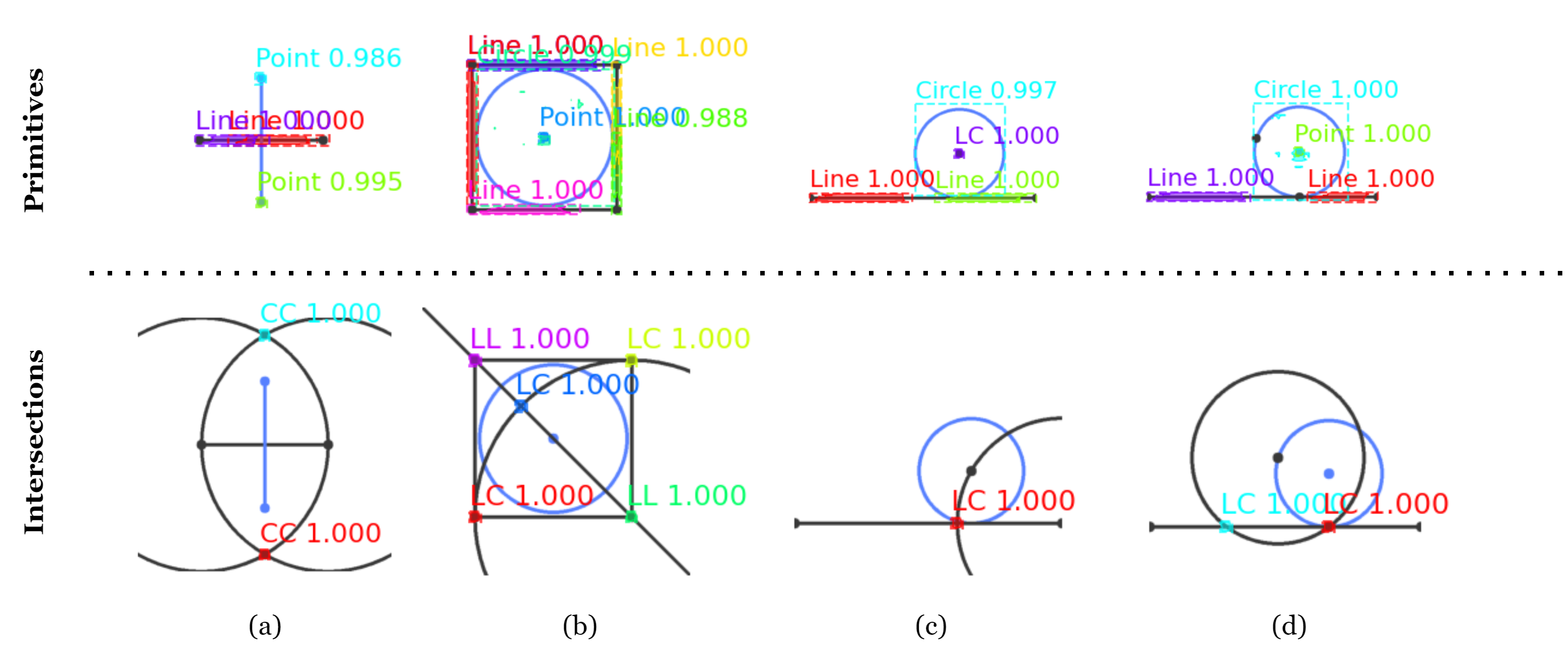}
  \caption{Illustration of segmentation for primitives and intersections. (a) Perpendicular Bisector (b) Circle in Square (c) Circle Tangent to Line (d) Circle through a Point Tangent to Line}
  \label{fig-maskrcnn_result}
\end{figure}

Table \ref{table-detection} shows the performance of the trained Mask R-CNN with different backbones. The result shows that our models can detect the targets with mAP over $90$\%. The illustration of the detection results for Euclidea problems is shown in Figure \ref{fig-maskrcnn_result}. Euclidea does not yet support questions involving the area, but it works well on the other constructible problems. We show the accuracy of solving geometric construction problems on the first six levels of Euclidea in Table \ref{table-evaluation}.

\begin{table}[!h]
\setlength\extrarowheight{0.5pt}
  \caption{PERFORMANCE COMPARISON: MASK R-CNN WITH BACKBONES RESNET-50 AND RESNET-101 ON THE PRIMITIVES AND INTERSECTION SEGMENTATION.}
  \label{table-detection}
  \centering
  \begin{tabular}{l|l|l l l || l|l l l}
    \toprule
   \multirow{2}{1em}{\textbf{Backbone}} & \multicolumn{4}{c||}{\textbf{Primitives}} &
    \multicolumn{4}{c}{\textbf{Intersections}}                   \\
         & mAP     & AP$_{50}$ & AP$_{75}$ & AP$_{90}$ & mAP & AP$_{50}$ & AP$_{75}$ & AP$_{90}$ \\
    \midrule
    ResNet-50 & 0.936  &  0.949 & 0.945 & 0.868 & 0.926  &  0.911 & 0.908 & 0.901\\
    ResNet-101 & 0.938  & 0.951 & 0.948 & 0.866 & 0.928  & 0.914 & 0.912 & 0.906\\
    \bottomrule
  \end{tabular}
\end{table}

\begin{table}[!h]
\caption{EVALUATION RESULT ON THE EUCLIDEA DATA SET}
\setlength\extrarowheight{0.5pt}
  \centering
\begin{tabular}{l|llllll}
\toprule
&\textbf{Alpha}&\textbf{Beta}&\textbf{Gamma}&\textbf{Delta}&\textbf{Epsilon}&\textbf{Zeta} \\
\midrule
\textbf{Accuracy} & 0.857 & 0.900 & 0.778 & 0.636 & 0.727& 0.636\\
\bottomrule
\end{tabular}
\label{table-evaluation}

\end{table}

\clearpage
\newpage
\section{Computational Examples} \label{appendixd}
\subsection{Euclidea Puzzle (\textit{Alpha-4}): Inscribed Circle}
Given a triangle, an inscribed circle is the largest circle contained within the triangle. The inscribed circle will touch each of the three sides of the triangle at exactly one point.

\begin{figure}[!h]
  \centering
  \includegraphics[width=\linewidth]{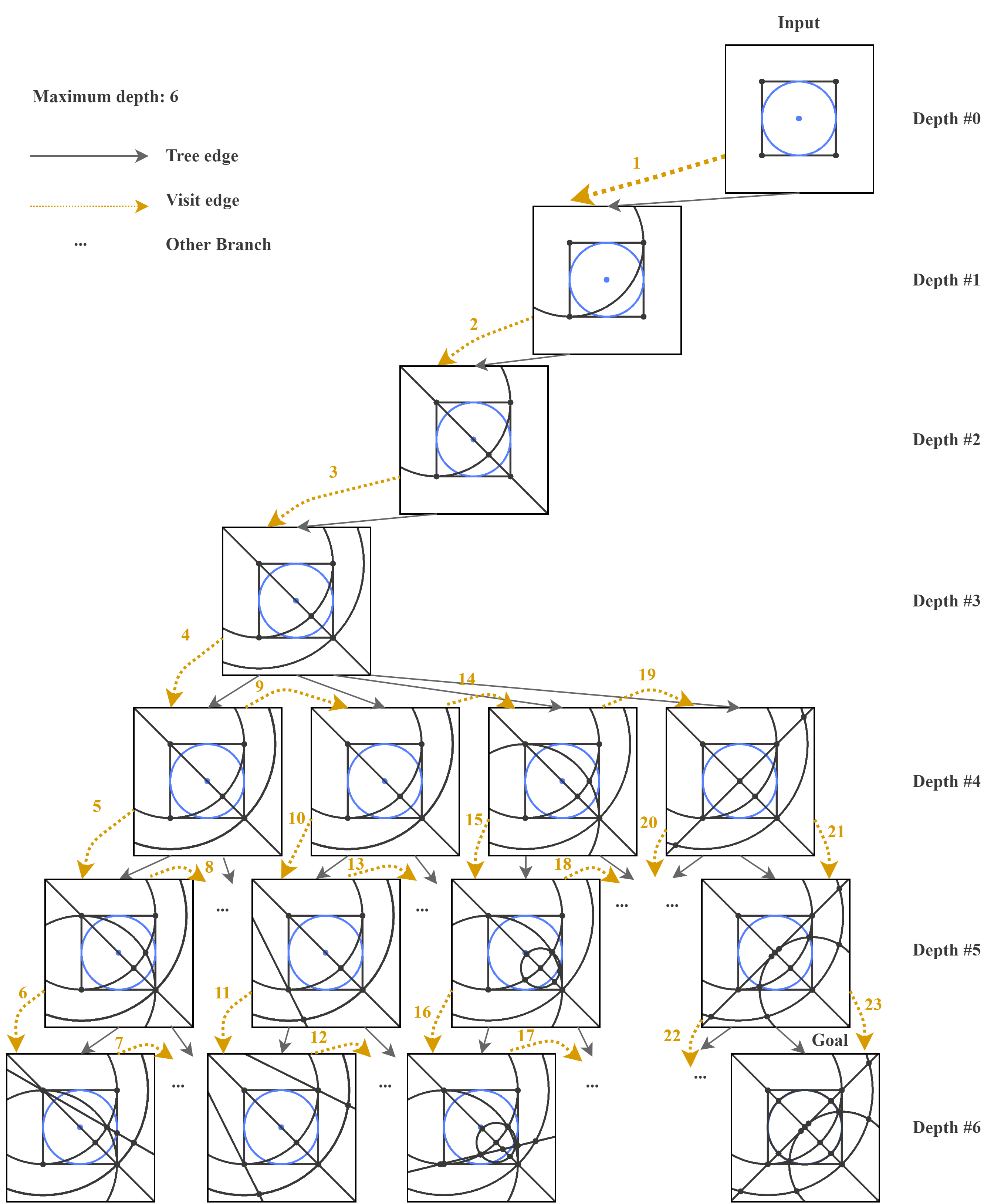}
  \caption{Illustration of the construction of a circle inscribed in the square (Euclidea \textit{Alpha-4}). The maximum depth in this example is 6.}
\end{figure}

\newpage
\subsection{Euclidea Puzzle (\textit{Gamma-5}): Circle Through a Point Tangent to Line}
A circle through a point tangent to the line is to construct a circle through the arbitrary point that is tangent to the given line at the point on the line.
\begin{figure}[!h]
  \centering
  \includegraphics[width=\linewidth]{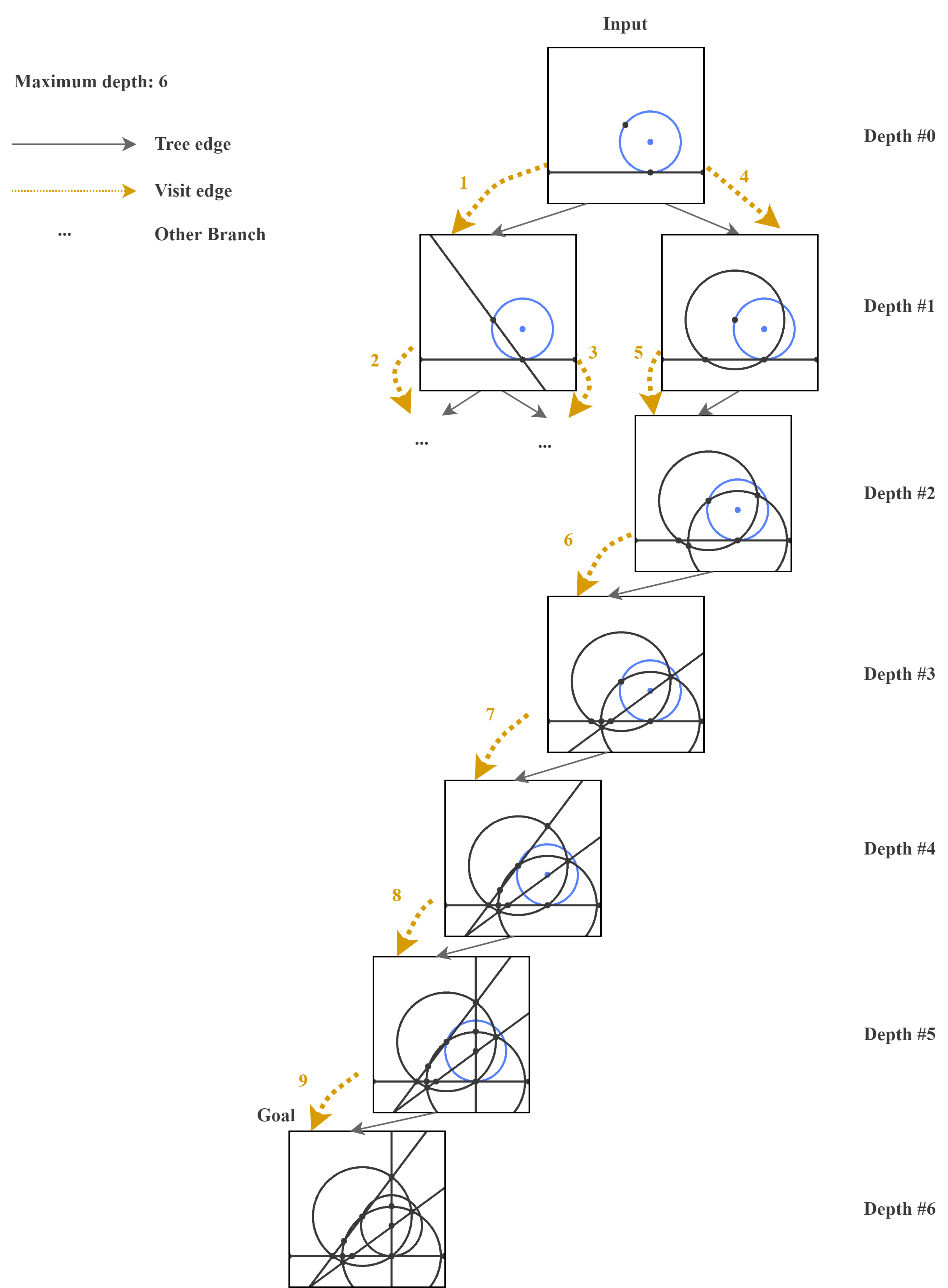}
  \caption{Illustration of the construction of a circle through a point tangent to line. The maximum depth in this example is 6.}
\end{figure}

\newpage
\subsection{Sangaku: Square and Circle in a Gothic Cupola}
Square and Circle in a Gothic Cupola from \cite{sangaku} are a Sangaku with two-quarter circles inscribed in a square form a gothic cupola. Inscribed in the latter is a circle on top, which stands a small circle.

\begin{figure}[!h]
  \centering
  \includegraphics[width=0.9\linewidth]{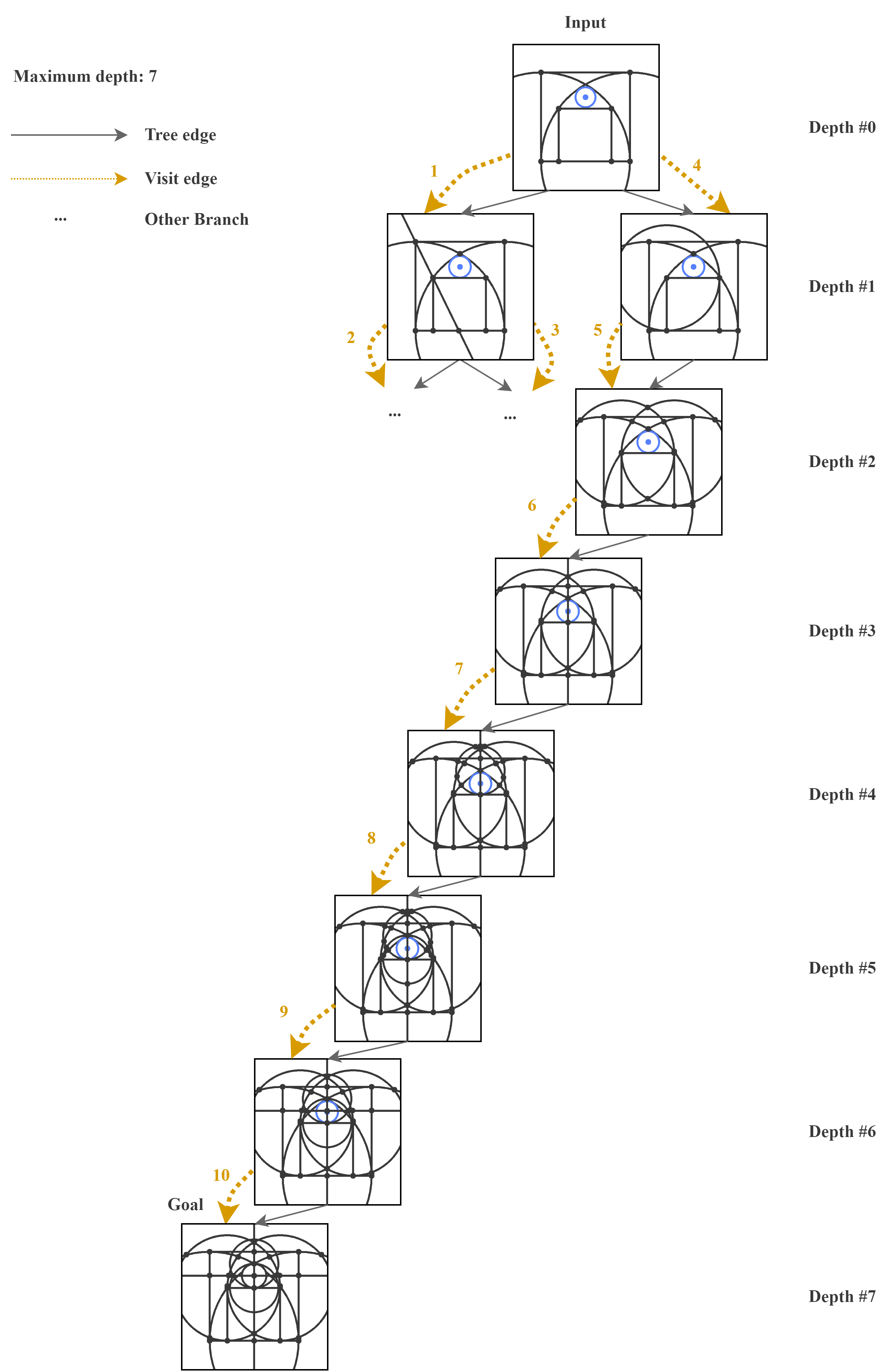}
  \caption{Illustration of the construction of a square and circle in a Gothic Cupola. The maximum depth in this example is 7.}
\end{figure}

\newpage
\subsection{Sangaku: Sangaku with Versines} Sangaku with Versines was written in 1825 in the Tokyo prefecture from \cite{sangaku}. It relates to the rarely used nowadays versine quantities and the distance from a vertex to the incircle.

\begin{figure}[!h]
  \centering
  \includegraphics[width=\linewidth]{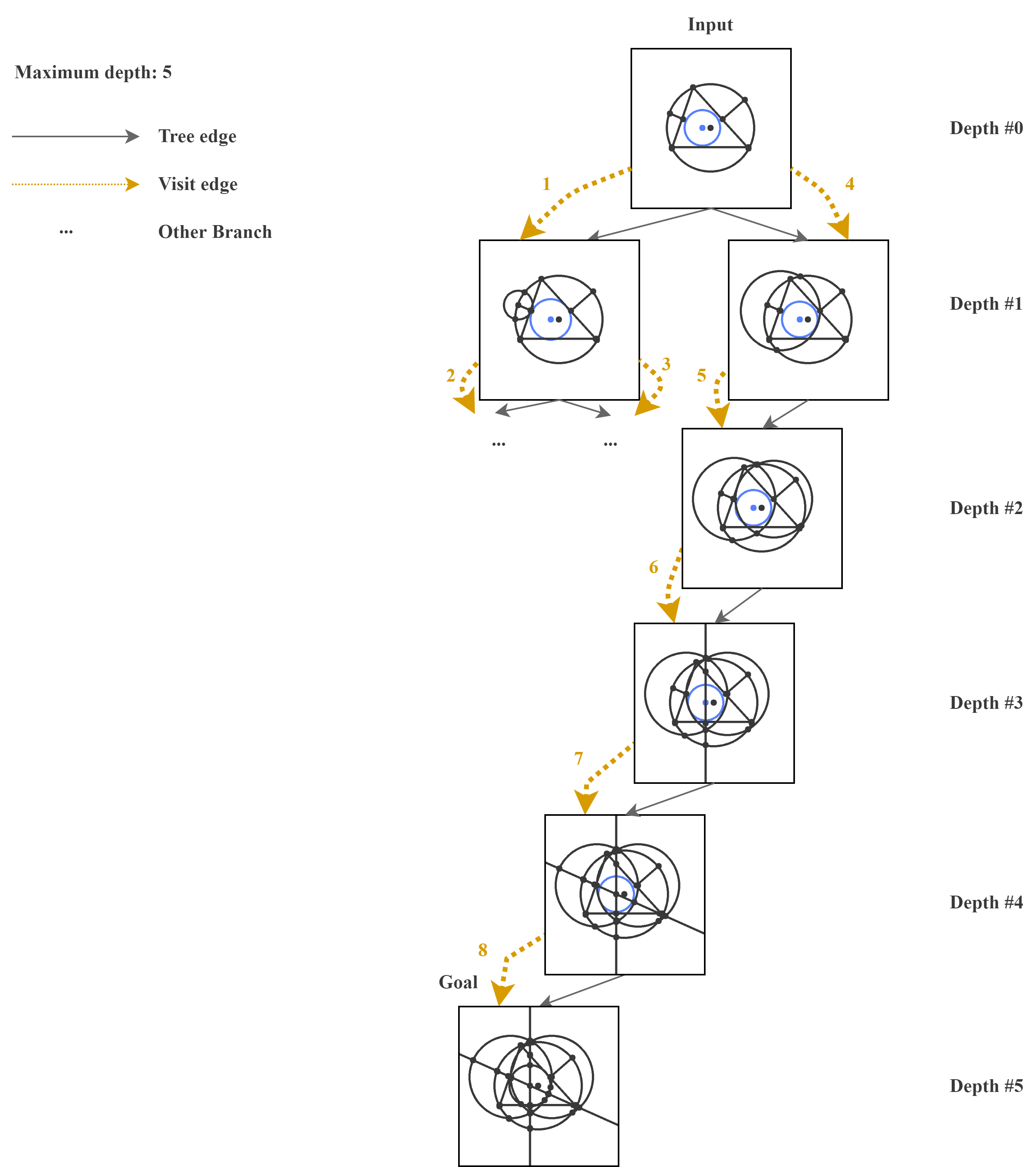}
  \caption{Illustration of the construction of Sangaku with versines. The maximum depth in this example is 5.}
\end{figure}

\end{document}